\documentclass[11pt,a4paper]{article}
\usepackage[hyperref]{naaclhlt2018}
\usepackage{times}
\usepackage{latexsym}
\usepackage{graphicx}
\usepackage{url}

\aclfinalcopy 

\title{NIHRIO at SemEval-2018 Task 3: A Simple and Accurate Neural Network Model for Irony Detection in Twitter}

\author{Thanh Vu$^1$, Dat Quoc Nguyen$^2$, Xuan-Son Vu$^3$, Dai Quoc Nguyen$^4$, \\ \textbf{Michael Catt}$^1$ \and \textbf{Michael Trenell}$^1$\\
$^1$NIHRIO, Newcastle University, UK; $^2$The University of Melbourne, Australia;\\
$^3$Ume\r{a} University, Sweden; $^4$Deakin University, Australia \\
{\tt thanh.vu@io.nihr.ac.uk; dqnguyen@unimelb.edu.au}; \\ {\tt sonvx@cs.umu.se}; {\tt dai.nguyen@deakin.edu.au};\\
{\tt \{michael.catt, michael.trenell\}@io.nihr.ac.uk}}

\date{}

\begin{document}
\maketitle
\begin{abstract}
This paper describes our NIHRIO system for SemEval-2018 Task 3 ``Irony detection in English tweets.'' We propose to use a simple neural network architecture of Multilayer Perceptron with various types of input features including: lexical, syntactic, semantic and polarity features.  Our system achieves very high performance in both subtasks of binary and multi-class irony detection in tweets. In particular, we rank \underline{third} using the {accuracy} metric and \underline{fifth} using the $F_1$ metric. Our code is available at: \url{https://github.com/NIHRIO/IronyDetectionInTwitter}.
\end{abstract}

\section{Introduction}

Mining Twitter data has increasingly been attracting much research attention in many NLP applications such as in sentiment analysis \cite{pak2010twitter, kouloumpis2011twitter, agarwal2011sentiment, liu2012emoticon, rosenthal2017semeval, cambria2018senticnet} and stock market prediction \cite{bollen2011twitter, vu2012experiment, bartov2015can, nofer2015using, oliveira2017impact}. Recently, \newcite{Davidov2010} and \newcite{Reyes2013} have shown that Twitter data includes a high volume of ``\emph{ironic}'' tweets. For example, a user can use positive words in a Twitter message to her intended negative meaning (e.g., \emph{``It is awesome to go to bed at 3 am \#not''}). This especially results in a research challenge to assign correct sentiment labels for ironic tweets \cite{bosco2013developing, ghosh2015semeval, Farias2016, nozza2017multi,kannangara2018mining}.

To handle that problem, much attention has been focused on automatic irony detection in Twitter \cite{Davidov2010, Reyes2013, barbieri2014modelling, rajadesingan2015sarcasm, Farias2016, sulis2016figurative, karoui2017exploring, joshi2017automatic, huang2017irony, ravi2017novel}. In this paper, we propose a neural network model for irony detection in tweets. Our model obtains the fifth best performances in both binary and multi-class irony detection subtasks in terms of $F_1$ score \cite{cynthia2018semeval}. Details of the two subtasks can be found in the task description paper  \citep{cynthia2018semeval}. We briefly describe the subtasks as follows:

\paragraph{Subtask 1 (A): Ironic vs non-ironic} This first subtask is a binary classification problem, in which we predict whether or not a tweet is \emph{ironic}. For example, \emph{``I just love when you test my patience!! \#not''} is ironic, but \emph{``Had no sleep and have got school now \#not happy''} is non-ironic.

\paragraph{Subtask 2 (B): Different types of irony} This second subtask is a multi-class classification problem, where we predict the correct label of a tweet from four classes: (1) \emph{non-irony},  (2) \emph{verbal irony by means of a polarity contrast},  (3) \emph{other verbal irony} and  (4) \emph{situational irony}.

\medskip

The remainder of this paper is organized as follows: We describe the ironic tweet dataset provided by the SemEval-2018 Task 3 in Section \ref{sec:data}. We then describe our system in Section \ref{sec:approach}. The experimental results and conclusion are detailed in Section \ref{sec:exp} and Section \ref{sec:concl}, respectively. 

\begin{figure*}[!ht]
\centering
\includegraphics[width=12cm]{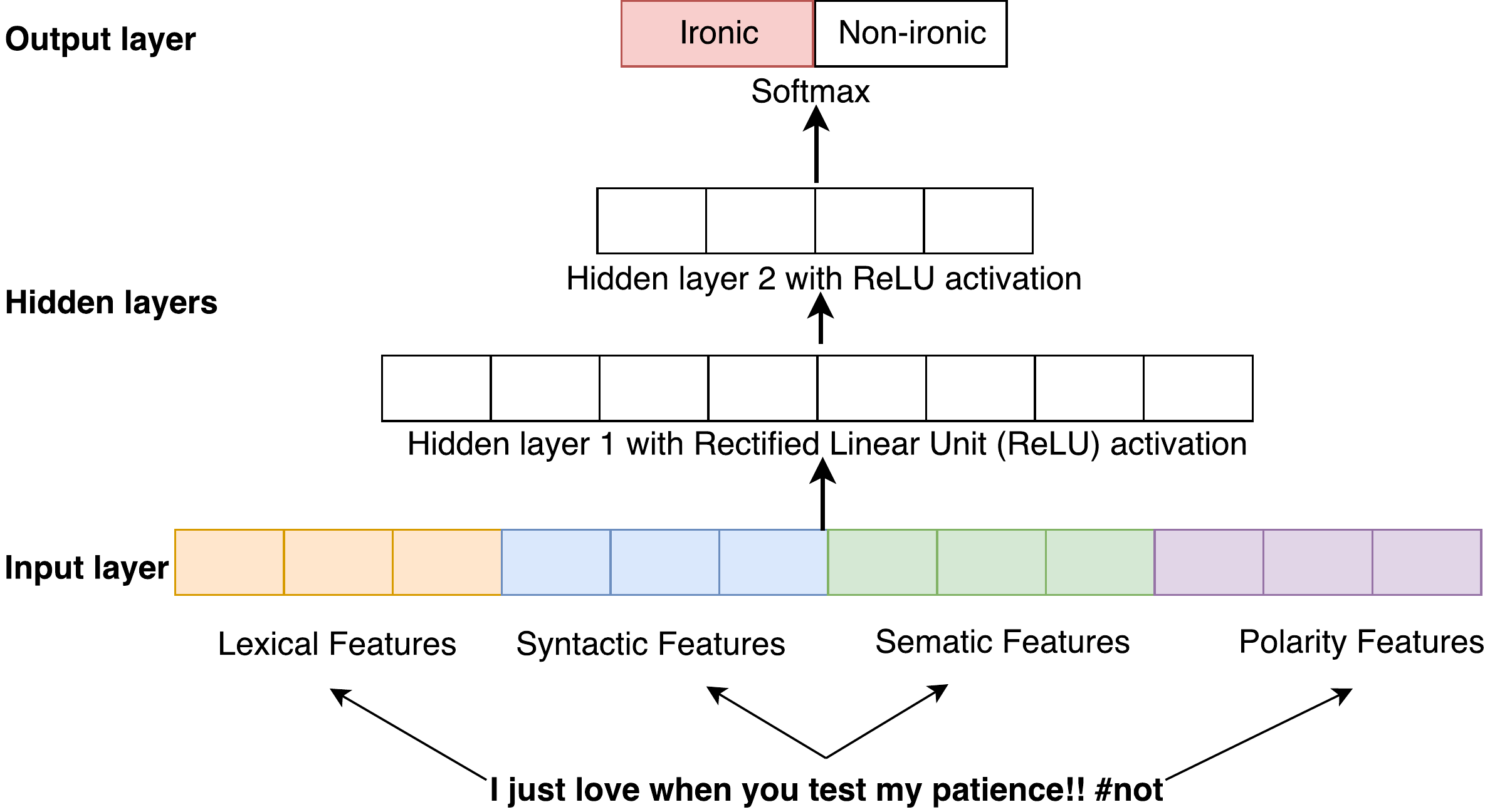} 
\caption{Overview of our model architecture for irony detection in tweets.}
\label{fig:diagram}
\end{figure*}

\section{Dataset}
\label{sec:data}

The dataset consists of 4,618 tweets (2,222 ironic + 2,396 non-ironic)  that are manually labelled by three students. Some pre-processing steps were applied to the dataset, such as the emoji icons in a tweet are replaced by a describing text using the Python emoji package.\footnote{\url{https://pypi.python.org/pypi/emoji/}} Additionally, all the ironic hashtags, such as \#not, \#sarcasm, \#irony, in the dataset have been removed. This makes difficult to correctly predict the label of a tweet. For example, \emph{``@coreybking thanks for the spoiler!!!! \#not''} is an ironic tweet but without \#not, it probably is a non-ironic tweet.  The dataset is split into the training and test sets as detailed in Table \ref{tab:tb1}.

\begin{table}[!t]
    \centering
    
    \begin{tabular}{l l l}
        \hline
        \textbf{Statistics} & \textbf{Training} & \textbf{Test}\\
        \hline
        \#samples & 3,834 & 784 \\
        \#non-irony & 1,923 & 473 \\
        \#irony & 1,911 & 311\\
        - \emph{polarity contrast verbal} & 1,390 & 164\\
        - \emph{other verbal} & 316 & 85\\
        - \emph{situational} & 205 & 62 \\\hline

    \end{tabular}
    \caption{Basic statistics of the provided dataset.}
    \label{tab:tb1}
\end{table}

Note that there is also an extended version of the training set, which contains the ironic hashtags. However, we only use the training set which does not contain the ironic hashtags to train our model as it is in line with the test set.

\paragraph{Our data pre-processing step:} 
Tweet normalization is an important pre-processing step as there are around 15\% of tweets containing 50\% or more out-of-vocabulary tokens \cite{han2011}. We normalize each tweet from the dataset using a lexicon-based approach proposed by \citet{han2012}, using a manually constructed normalization dictionary (e.g., ``reeeaaalll'' is normalized by ``real'). We then replace all tagged users and urls by specific word tokens ``$<$USER$>$'' and ``$<$URL$>$'', respectively. It is because they are likely not correlated with the ironic labels.

\section{Our modeling approach}
\label{sec:approach}
 
 We first describe our MLP-based model for ironic tweet detection in Section \ref{sec:model}. We then present the features used in our model in Section \ref{sec:feature}.

\subsection{Neural network model}
\label{sec:model}

We propose to use the Multilayer Perceptron (MLP) model \cite{Hornik1989} to handle both the ironic tweet detection subtasks. Figure \ref{fig:diagram} presents an overview of our model architecture including an input layer, two hidden layers and a softmax output layer.  Given a tweet, the input layer represents the tweet by a feature vector which concatenates lexical, syntactic, semantic and polarity feature representations.  The two hidden layers with ReLU activation function take the input feature vector to select the most important features which are then fed into the softmax layer for ironic detection and classification.

\subsection{Features}
\label{sec:feature}

Table \ref{tab:tb2} shows the number of lexical, syntactic, semantic and polarity features used in our model.

\paragraph{Lexical features:}
Our lexical features include 1-, 2-, and 3-grams in both word and character levels. For each type of $n$-grams, we utilize only the top 1,000 $n$-grams based on the term frequency-inverse document frequency (tf-idf) values. That is, each $n$-gram appearing in a tweet becomes an entry in the feature vector with the corresponding feature value tf-idf. We also use the number of characters and the number of words as features. 

\paragraph{Syntactic features:}
We use the NLTK toolkit to tokenize and annotate part-of-speech tags (POS tags) for all tweets in the dataset. We then use all the POS tags with their corresponding tf-idf values as our syntactic features and feature values, respectively. 

\begin{table}[!t]
    \centering
    
    \begin{tabular}{l l l}
        \hline
        \textbf{Name} & \textbf{\# Features} \\
        \hline
        Lexical features & 2,002 \\
        Syntactic features & 45 \\
        Semantic features & 700\\
        Polarity features & 12 \\
       \hline
       \textbf{Total} & \textbf{2,759} \\\hline
    
    \end{tabular}
    \caption{Number of features used in our model}
    \label{tab:tb2}
\end{table}

\paragraph{Semantic features:}
A major challenge when dealing with the tweet data is that the lexicon used in a tweet is informal and much different from tweet to tweet. The lexical and syntactic features seem not to well-capture that property. To handle this problem, we apply three approaches to compute tweet vector representations. 

\begin{table}[!t]
    \centering
    
    \begin{tabular}{l l | l l}
        \hline
        \textbf{Cluster} & \textbf{Word} & \textbf{Cluster} & \textbf{Word}\\
        \hline
       110000 & wife & 11001000    & adorable\\
       110000 & sister & 11001000 & excellent\\
       110000 & boyfriend & 11001000 & interesting\\
       110000 & daughter & 11001000 & blessed \\
       110000 & mum & 11001000 & easy\\
       110000 & son & 11001000 & perfect\\
       110000 & dad & 11001000 & cool\\
       110000 & family & 11001000 & funny\\
       \hline
    \end{tabular}
    \caption{Example of clusters produced by the Brown clustering algorithm.}
    \label{tab:tbcluster}
\end{table}

Firstly, we employ 300-dimensional pre-trained word embeddings from GloVe \cite{pennington2014glove} to compute a tweet embedding as the average of the embeddings of words in the tweet. 

Secondly, we apply the latent semantic indexing  \cite{Papadimitriou1998} to capture the underlying semantics of the dataset. Here, each tweet is represented as a vector of 100 dimensions.

Thirdly, we also extract tweet representation by applying the Brown clustering algorithm \cite{Brown1992, liang2005semi}\footnote{\url{https://github.com/percyliang/brown-cluster}}---a hierarchical clustering algorithm which groups the words with similar meaning and syntactical function together. Applying the Brown clustering algorithm, we obtain a set of clusters, where each word belongs to only one cluster. For example in Table \ref{tab:tbcluster}, words that indicate the members of a family (e.g., ``mum'', ``dad'') or positive sentiment (e.g., ``interesting'', ``awesome'') are grouped into the same cluster. 
We run the algorithm with different number of clustering settings (i.e., 80, 100, 120) to capture multiple semantic and syntactic aspects. For each clustering setting, we use the number of tweet words in each cluster as a feature. After that, for each tweet, we concatenate the features from all the clustering settings to form a cluster-based tweet embedding.

\paragraph{Polarity features:}
Motivated by the verbal irony by means of polarity contrast, such as \emph{``I really love this year's summer; weeks and weeks of awful weather''},  we use the number of polarity signals appearing in a tweet as the polarity features. The signals include positive words (e.g., love), negative words (e.g., awful), positive emoji icon and negative emoji icon. We use the sentiment dictionaries provided by \citet{hu2004mining} to identify positive and negative words in a tweet. We further use boolean features that check whether or not a negation word is in a tweet (e.g., \emph{not}, \emph{n't}).

\begin{figure}[!t]
\centering
\includegraphics[width=4.5cm]{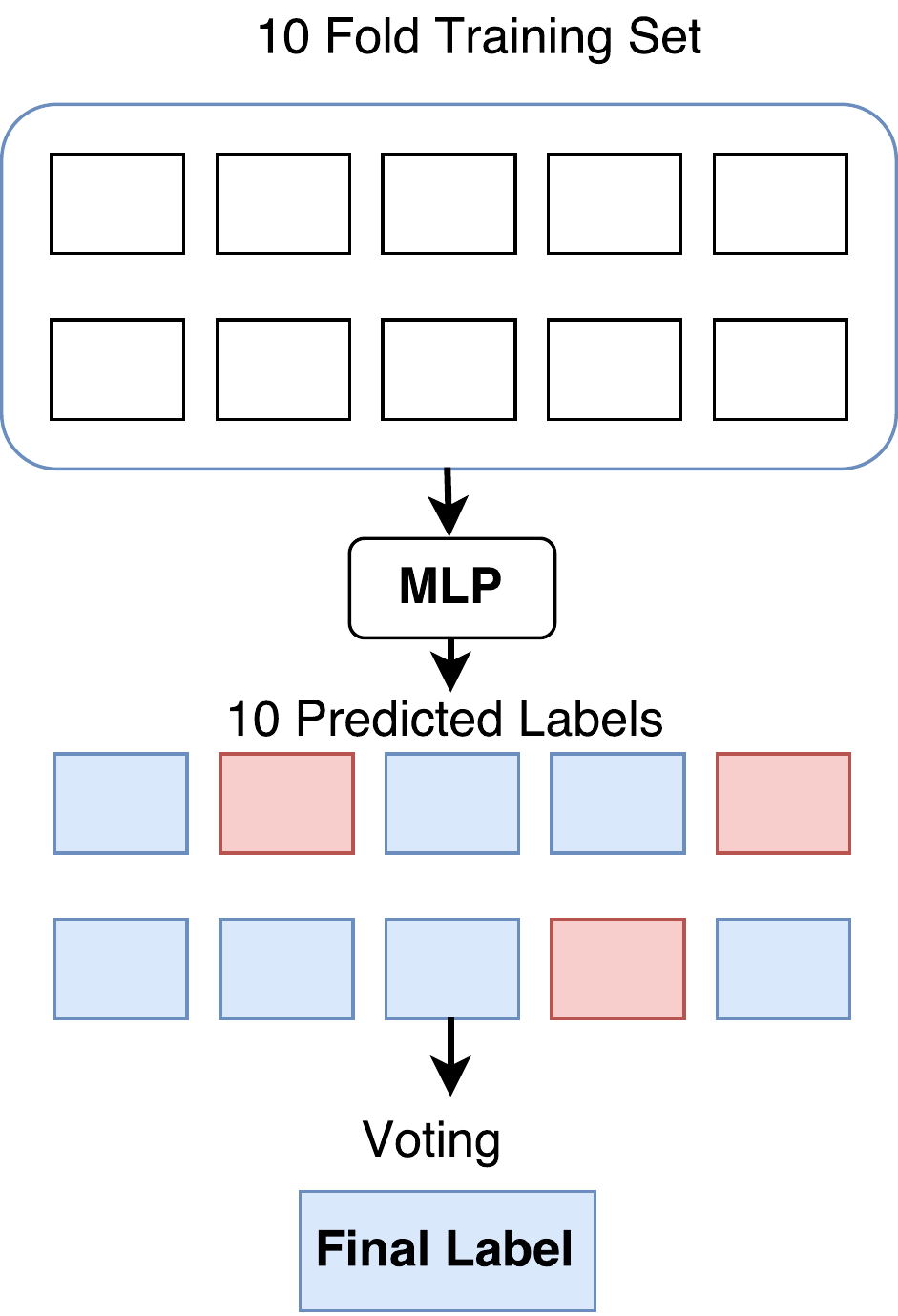} 
\caption{The training mechanism.}
\label{fig:training}
\end{figure}

\subsection{Implementation details}
We use Tensorflow \cite{tensorflow2015-whitepaper} to implement our model. Model parameters are learned to minimize the  the cross-entropy loss
with L$_2$ regularization.  
Figure \ref{fig:training} shows our training mechanism. In particular, we follow a $10$-fold cross-validation based voting strategy.  First, we split the training set into 10 folds. Each time, we combine 9 folds to train a  classification model and use the remaining fold to find the optimal hyperparameters.  Table \ref{tab:tb3} shows optimal settings for each subtask. 

In total, we have 10 classification models to produce 10 predicted labels for each test tweet. Then, we use the voting technique to return the final predicted label.

\begin{table}[!t]
    \centering
    
    \begin{tabular}{l l l}
        \hline
        \textbf{Name} & \textbf{1 (A)} & \textbf{ 2 (B)} \\
        \hline
        Hidden layers & (800, 400) & (800, 300) \\
        \# epoch & 100 & 100 \\
        early stop & 30 & 30 \\
        Learning rate & 10$^{-4}$& 10$^{-4}$\\
        $l_2$ & 10$^{-5}$ & 10$^{-5}$ \\\hline
    
    \end{tabular}
    \caption{The optimal hyperparameter settings for subtasks 1 (A) and 2 (B).}
    \label{tab:tb3}
\end{table}

\section{Experiments}
\label{sec:exp}

\subsection{Metrics}
The metrics used to evaluate our model include \emph{accuracy}, \emph{precision}, \emph{recall} and \emph{F$_1$}. The accuracy is calculated using all classes in both tasks. The remainders are calculated using only the positive label in subtask 1 or per class label (i.e., macro-averaged) in subtask 2. Detail description of the metrics can be found in \citet{cynthia2018semeval}.

\subsection{Results for subtask 1}
Table \ref{tab:result1} shows our official results on the test set for subtask 1 with regards to the four metrics. By using a simple MLP neural network architecture, our system achieves a high performance which is ranked \textbf{third} and \textbf{fifth} out of forty-four teams using accuracy and \emph{F$_1$}  metrics, respectively. 

\begin{table}[!t]
    \centering
    
    \begin{tabular}{l l l l l}
        \hline
       \textbf{Accuracy} & \textbf{Precision} & \textbf{Recall} & \textbf{F$_1$}\\\hline
       70.15\textsubscript{3} & 60.91 & 69.13 & 64.76\textsubscript{5}
     
       \\\hline
    
    \end{tabular}
    \caption{The performance (in \%) of our model on the test set for subtask 1 (binary classification). The subscripts denote our official ranking.}
    \label{tab:result1}
\end{table}

\subsection{Results for subtask 2}
Table \ref{tab:result2} presents our results on the test set for subtask 2. Our system also achieves a high performance which is ranked \textbf{third} and \textbf{fifth} out of thirty-two teams using accuracy and \emph{F$_1$}  metrics, respectively. We also show in Table \ref{tab:result3}  the performance of our system on different class labels. For ironic classes, our system achieves the best performance on the verbal irony by means of a polarity contrast with $F_1$ of 60.73\%. Note that the performance on the situational class is not high. The reason is probably that the number of situational tweets in the training set is small (205/3,834), i.e. not enough to learn a good classifier.

\begin{table}[!t]
    \centering
    
    \begin{tabular}{l l l l l}
        \hline
       \textbf{Accuracy} & \textbf{Precision} & \textbf{Recall} & \textbf{F$_1$}\\\hline
       65.94\textsubscript{3} & 54.46 & 44.75 & 44.37\textsubscript{5}
     
       \\\hline
    
    \end{tabular}
    \caption{The performance (in \%) of our model on the test set for subtask 2 (multi-class classification).}
    \label{tab:result2}
\end{table}

\begin{table}[!t]
    \centering
    
    \begin{tabular}{l l l l l}
        \hline
       \textbf{Class} & \textbf{Precision} & \textbf{Recall} & \textbf{F$_1$}\\\hline
       Non-irony & 72.97 & 79.92 & 76.29 \\
       Contrast verbal & 53.21 & 70.73 & 60.73\\
       Other verbal & 48.78 & 23.53 & 31.75\\
       Situational & 42.86 & 4.84 & 8.70\\
     
       \hline
    
    \end{tabular}
    \caption{The performance (in \%) of our model on the test set for each class label in subtask 2.}
    \label{tab:result3}
\end{table}

\subsection{Discussions}
Apart from the described MLP models, we have also tried other neural network models, such as Long Short-Term Memory (LSTM) \cite{Hochreiter1997} and Convolutional Neural Network (CNN) for relation classification  \cite{kim2014convolutional}. 
We found that LSTM achieves much higher performance than MLP does on the extended training set containing the ironic hashtags (about 92\% vs 87\% with 10-fold cross-validation using $F_1$ on subtask 1). However, without the ironic hashtags, the performance is lower than MLP's. We also employed popular machine learning techniques, such as SVM \cite{hearst1998support}, Logistic Regression \cite{harrell2001ordinal}, Ridge Regression Classifier \cite{le1992ridge}, but none of them produces as good results as MLP does.  We have also implemented ensemble models, such as voting, bagging and stacking. We found that with 10-fold cross-validation based voting strategy, our MLP models produce the best irony detection and classification results.   

\section{Conclusion}
We have presented our NIHRIO system for participating the Semeval-2018 Task 3 on ``Irony detection in English tweets''. We proposed to use Multilayer Perceptron to handle the task using various features including lexical features, syntactic features, semantic features and polarity features. Our system was ranked the fifth best performing one with regards to \emph{F$_1$} score in both the subtasks of binary and multi-class irony detection in tweets. 
\label{sec:concl}

\section*{Acknowledgments}
 This research is supported by the  National Institute for Health Research (NIHR) Innovation Observatory at Newcastle University, United Kingdom.
 
\bibliography{refs}
\bibliographystyle{acl_natbib}
\end{document}